\documentclass[runningheads]{llncs}

\usepackage{eccv}

\usepackage{eccvabbrv}

\usepackage{graphicx}
\usepackage{booktabs}

\usepackage[accsupp]{axessibility}  
\usepackage[table]{xcolor}
\usepackage{multirow}
\usepackage[utf8]{inputenc}
\usepackage[most]{tcolorbox}
\usepackage{makecell}

\usepackage{hyperref}

\usepackage{orcidlink}

\begin{document}

\title{PatchCue: Enhancing Vision-Language Model Reasoning with Patch-Based Visual Cues} 

\titlerunning{PatchCue}


\author{
Yukun Qi \and
Pei Fu \and
Hang Li \and
Yuhan Liu \and
Chao Jiang \and
Bin Qin \and
Zhenbo Luo\thanks{Corresponding author} \and
Jian Luan
}

\authorrunning{Y. Qi et al.}

\institute{
MiLM Plus, Xiaomi Inc.\\
\email{\{qiyukun, luozhenbo\}@xiaomi.com}
}


\maketitle

\begin{abstract}
Vision-Language Models (VLMs) have achieved remarkable progress on a wide range of challenging multimodal understanding and reasoning tasks. However, existing reasoning paradigms, such as the classical Chain-of-Thought (CoT), rely solely on textual information and often underutilize important visual cues. While prior work has incorporated pixel-level visual cues, these representations require precise spatial localization, introducing additional learning complexity. To address this, we propose PatchCue, a novel patch-based visual cue paradigm designed to significantly enhance the visual reasoning capabilities of VLMs. By partitioning images into patches and representing cues at the patch level, PatchCue aligns better with human perceptual habits and leverages the patch-tokenized input of modern VLMs. We train VLMs using a two-stage approach: cold-start supervised fine-tuning to output patch-level cues, followed by reinforcement learning with a process-supervised cue reward that guides intermediate visual reasoning steps. Extensive experiments on multiple VLMs and diverse benchmarks, including general visual question answering, complex reasoning, and document understanding, demonstrate that PatchCue consistently improves overall model performance. Our results show that patch-level cues outperform both pixel-level bounding boxes and point-based cues, providing a more effective and cognitively aligned visual reasoning paradigm.

\end{abstract}
    
\section{Introduction}

In recent years, Vision-Language Models (VLMs) have made remarkable progress across a wide range of multimodal understanding and reasoning tasks \cite{hurst2024gpt, comanici2025gemini, bai2025qwen2, coreteam2025mimovltechnicalreport, seed2025seed1_5vl}. 
As tasks grow more complex, recent studies highlight the importance of \textit{thinking with images}—reasoning that repeatedly consults visual information rather than relying solely on text. 
This moves beyond the classical Chain-of-Thought (CoT) paradigm, which depends exclusively on textual reasoning \cite{wei2022chain, team2025kimi, deepseekai2025deepseekr1incentivizingreasoningcapability}, motivating approaches that incorporate visual cues into intermediate reasoning steps \cite{shao2024visual, zheng2025deepeyes, zhang2025thyme, qi2024cogcom}. 
Such interleaved visual-text reasoning improves both accuracy and interpretability.

Existing approaches can broadly be classified into two overarching categories: (1) \textbf{Externally-Guided Reasoning}, emulating how humans rely on external tools to inspect images \cite{zheng2025deepeyes, zhang2025thyme, liu2025visual, hu2024visual, su2025openthinkimg}. These methods train models to invoke tools such as object detectors, cropping modules, or magnifiers during the reasoning process, enabling them to isolate important regions and incorporate the resulting visual cues to support inference. (2) \textbf{Internally-Driven Reasoning}, aiming to activate the model’s intrinsic ability to explore visual cues. Instead of depending on external modules, these approaches prompt the model to repeatedly attend to the image throughout reasoning, progressively identifying and leveraging salient regions to enhance inference performance \cite{shao2024visual, qi2024cogcom, wang2025vgr, gao2025interleaved, yang2025look}.

Essentially, both types of approaches aim to identify key visual cue regions within an image and represent them in a form that effectively assists model reasoning. Currently, the dominant form of visual cue representation is at the pixel level, where critical regions are described by precise spatial coordinates \cite{zhang2025thyme, shao2024visual, chen2025sifthinker}. Such fine-grained representations require detailed visual perception capabilities and introduce additional learning complexity. From the perspective of human visual cognition, individuals often rely on approximate cue regions rather than precise coordinates when interpreting visual scenes. For example, when asked “Which person is speaking in the picture?”, humans tend to focus on the speaker’s head or mouth region without needing to pinpoint the exact pixel boundaries. This suggests that in many visual reasoning scenarios, coarse spatial localization is sufficient to support accurate inference. These observations naturally raise an intriguing question: \textit{Is there a more efficient and cognitively aligned form of visual cue representation that can better support multimodal reasoning?}

\begin{figure*}[t]
  \centering
  \includegraphics[width=1.\textwidth]{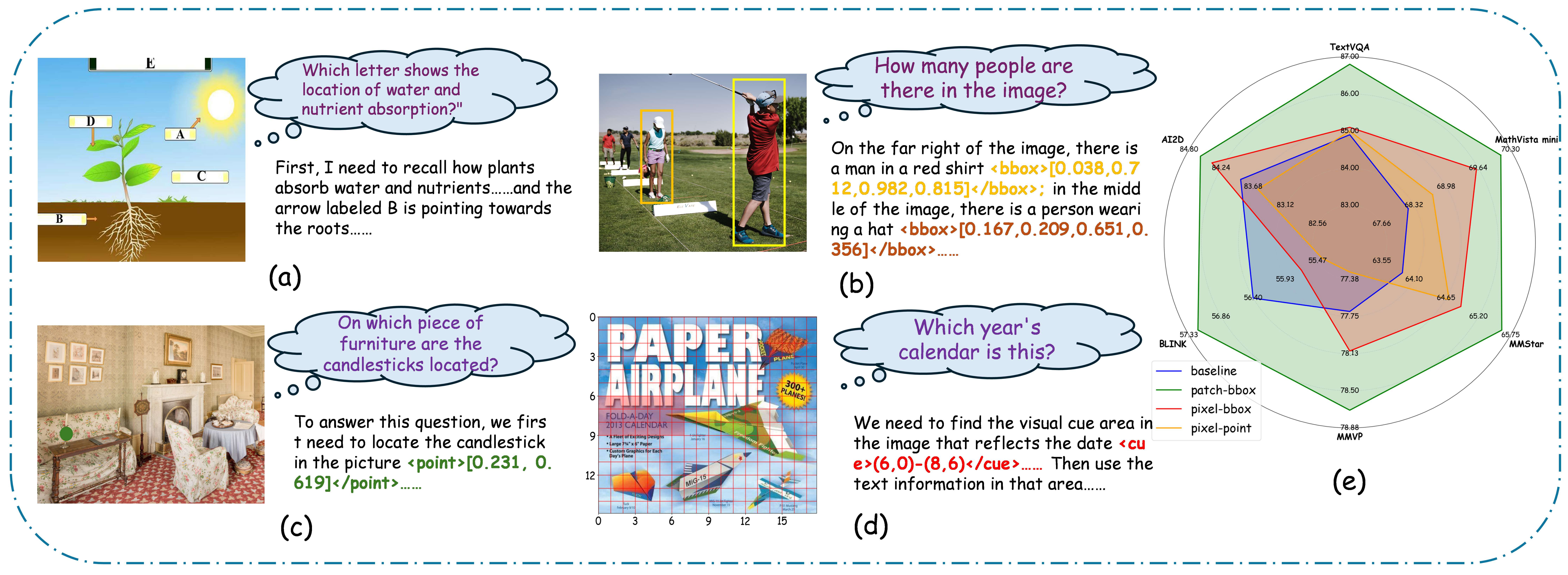}
  \caption{\textbf{Comparison of reasoning with different cue types:} (a) Text-only: reasoning based solely on textual information; (b) Pixel-bbox: cues represented as precise pixel-level bounding boxes; (c) Pixel-point: cues indicated by single pixel points highlighting key regions; (d) Patch-bbox: cues represented as patch-level regions to capture localized visual information; (e) SFT training comparison shows that patch-based cues improve model performance more effectively than pixel-bbox or pixel-point cues.} 
  \label{fig:main}
\end{figure*}

To investigate this question, we analyze several representative visual cue forms, as illustrated in Figure~\ref{fig:main}. The text-only paradigm reflects reasoning occurring purely in the mind after initial observation, without iterative interaction with visual information. Pixel-level cues are typically represented as pixel-bbox base \cite{shao2024visual, qi2024cogcom, wang2025vgr, chen2025sifthinker} or pixel-point base \cite{zhang2025hyperclick, wu2025gui, yang2025kwai}. While pixel-bbox cues require precise spatial localization, which may impose unnecessary granularity, point cues are simpler but convey limited and sometimes ambiguous information. Motivated by the patch tokenization mechanism in modern VLMs \cite{bai2025qwen2, coreteam2025mimovltechnicalreport, yang2025kwai}, we introduce a patch-bbox-based visual cue representation, partitioning the image into multiple patches and using patch coordinates to encode visual cues. As shown in Figure~\ref{fig:main} (e), validation experiments on Qwen2.5-VL-7B \cite{bai2025qwen2} show that, under the same data scale, patch-level cues outperform both pixel-bbox and pixel-point cues, highlighting their effectiveness in enhancing multimodal reasoning.

Building on these insights, we propose PatchCue, a patch-bbox visual cue paradigm designed to enhance the visual reasoning capabilities of VLMs. Using generated patch-cue data, models are trained in two stages: cold-start supervised fine-tuning (SFT) to produce patch-level cues, followed by Group Relative Policy Optimization (GRPO) \cite{shao2024deepseekmath} for reinforcement learning. Unlike standard GRPO, PatchCue supervises intermediate patch regions, enabling more controllable optimization. A cue reward encourages accurate and informative cues while preventing over-reliance, improving the coherence and interpretability of interleaved visual–text reasoning. Experiments across multiple benchmarks show that PatchCue consistently improves performance, e.g., yielding an average gain of 2 points on Qwen2.5-VL-7B \cite{bai2025qwen2}, demonstrating its effectiveness and strong generalization.

Our main contributions are as follows:
\begin{itemize}
    \item We propose a patch-bbox visual cue representation that partitions images into patches and encodes key regions with patch coordinates, improving multimodal reasoning efficiency and aligning better with human perception compared to pixel-level cues.
    \item By combining cold-start SFT with an improved GRPO, intermediate patch regions are explicitly supervised, and a cue reward guides the model to focus on informative visual cues for controllable visual–text reasoning.
    \item Experiments on multiple vision–language benchmarks with Qwen2.5-VL-7B \cite{bai2025qwen2} show that PatchCue consistently outperforms pixel-level cues, achieving an average improvement of 2 points and enhancing both accuracy and interpretability.
\end{itemize}

\section{Related Work}

\paragraph{\textbf{Thinking with Images.}}
With the rapid development of large language models (LLMs), vision-language models (VLMs) have emerged as powerful systems capable of complex multimodal reasoning, achieving significant progress in areas such as open-source model development \cite{liu2024llava, chen2024internvl, bai2025qwen2, huang2026vision}, dataset construction \cite{chen2024sharegpt4v, zeng2025enhancing, chen2024sharegpt4video}, evaluation protocols \cite{chen2024we, qi2025vcr, zhao2025v2p, zeng2026vision}, and novel training objectives and architectural designs \cite{fang2023eva, wang2023internimage, liu2025mind, liu2024semantic}. Unlike text-only reasoning, which treats visual information as a static initial context \cite{wei2022chain, team2025kimi, deepseekai2025deepseekr1incentivizingreasoningcapability}, the “thinking with images” paradigm actively leverages visual information as intermediate steps during the reasoning process, becoming a key focus in VLM research. Existing approaches can be broadly categorized into two types. The first relies on external tools for additional visual processing and interaction, such as Deepeyes \cite{zheng2025deepeyes}, VRAG-RL \cite{wang2025vrag}, Visual-ARFT \cite{liu2025visual}, and Thyme \cite{zhang2025thyme}. The second type exploits the model’s intrinsic capabilities, interleaving visual cues directly within the textual reasoning pipeline. Early works such as VisualCoT \cite{shao2024visual} and CogCom \cite{qi2024cogcom} primarily employ bounding boxes as visual hints, while more recent studies explore richer visual representations. For example, Look-Back \cite{yang2025look} uses text-visual prompting to trigger reflective reasoning, PaDT \cite{su2025patch} incorporates visual encoding-decoding modules to enhance visual grounding, and MINT-CoT \cite{chen2025mint} introduces patch-level visual cues in geometric reasoning tasks. These advances collectively provide important possibilities for developing more general and effective interleaved visual-text reasoning paradigms.

\paragraph{\textbf{Reinforcement Learning for Vision-Language Models.}}
Reinforcement learning (RL) \cite{schulman2017proximal, rafailov2023direct, shao2024deepseekmath} has been widely adopted to enhance the reasoning capabilities of language models, as demonstrated by the success of DeepSeek-R1 \cite{guo2025deepseek} in mathematical reasoning tasks. Building on these advances, recent studies have extended RL to VLMs, with rule-based RL in multimodal domains emerging as a particularly promising direction. For perception enhancement, R1-V \cite{chen2025r1v} applies RL to object counting, while Perception-R1 \cite{yu2025perception} leverages object matching and IoU as reward signals to improve visual grounding. In terms of reasoning, MMEureka \cite{meng2025mm} demonstrates the effectiveness of rule-based RL in mathematical problem-solving, and AGILE \cite{zeng2025agentic} enhances model reasoning through specialized visual tasks. From a data perspective, Vision-R1 \cite{huang2025vision} and R1-OneVision \cite{yang2025r1} convert visual information into textual representations to construct multimodal CoT datasets that facilitate stronger reasoning. Despite these significant advances, the complexity of the visual reasoning process still makes it highly challenging to apply RL supervision to intermediate reasoning steps, limiting the effectiveness of RL for fine-grained visual reasoning.
\section{Method}
\subsection{Overview}

\begin{figure*}[t]
  \centering
  \includegraphics[width=1.\textwidth]{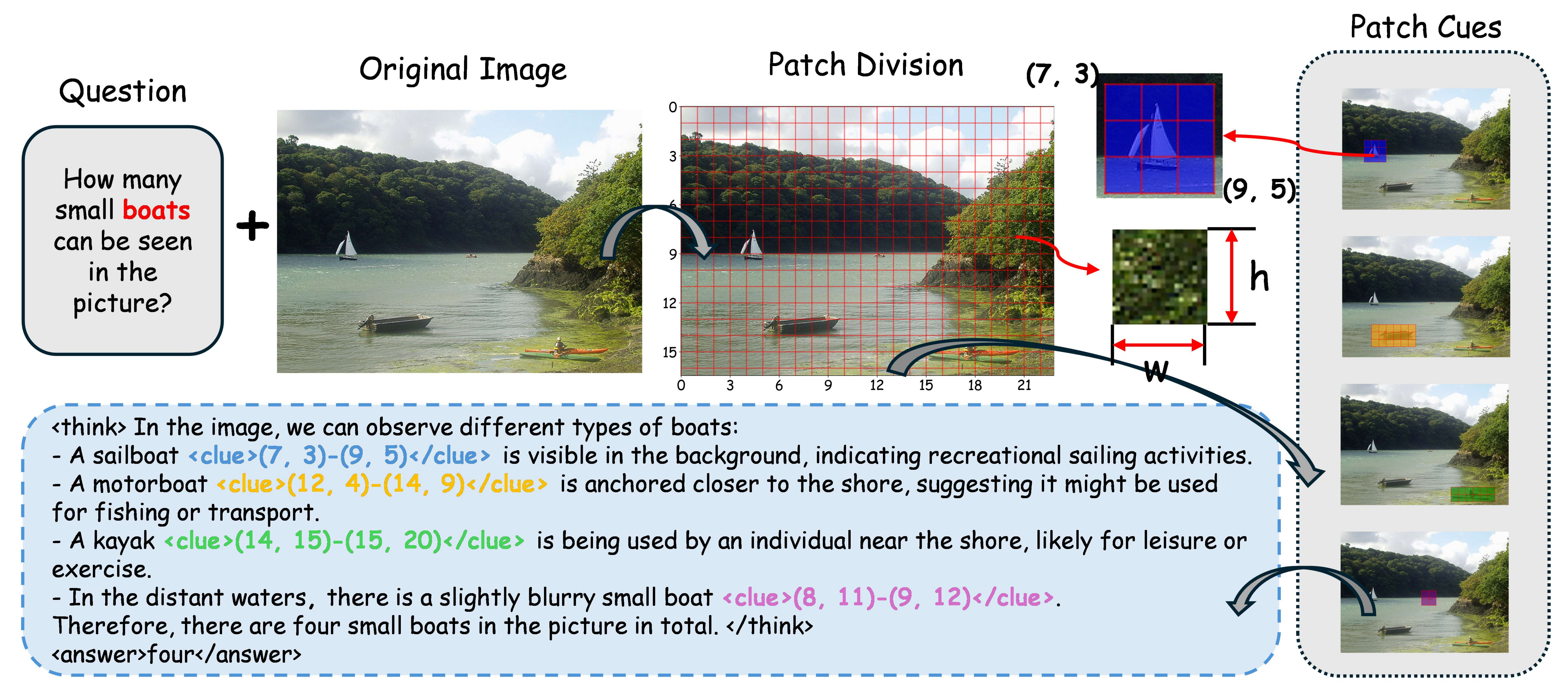}
  \caption{\textbf{Overview of PatchCue.} We divide images into fixed-size patches in order to represent important regions as visual cues. During the model’s reasoning process, it is essential not only to identify which patches are relevant to the given question but also to accurately reference and integrate these cues throughout each reasoning step. This structured use of patch-level cues helps the model ground its intermediate reasoning in the visual content, improving both interpretability and overall performance.} 
  \label{fig:case}
\end{figure*}

We propose PatchCue, a framework that enhances the reasoning capability of VLMs through patch-bbox visual cues. As illustrated in Figure~\ref{fig:case}, PatchCue introduces interpretable visual cues into the reasoning process, enabling dynamic interaction between textual reasoning and visual attention. This design allows the model to actively refer to visual evidence throughout reasoning, thereby improving both its visual sensitivity and overall reasoning consistency. In Section~\ref{sec:patch cues}, we define and formalize the concept and representation of patch-bbox visual cues. In Section~\ref{Data Construction}, we describe the visual cue data construction pipeline, which identifies key visual regions from multimodal datasets, generates high-quality patch-based cues, and reconstructs the corresponding reasoning trajectories. In Section~\ref{sec: traning}, we present the cue-based training paradigm and introduce a novel process-supervised learning approach that provides fine-grained rewards and constraints for cue generation during reasoning. This mechanism enables more controllable optimization of the intermediate reasoning process.

\subsection{Patch Cues}
\label{sec:patch cues}
Pixel-level visual cues are typically represented using either absolute or relative spatial coordinates. In the absolute case, a pixel-level bounding box is represented by its top-left and bottom-right coordinates $(x_1, y_1)$ and $(x_2, y_2)$, whereas in the relative case, coordinates are normalized to $[0,1]$ and must be scaled according to image dimensions $H$ and $W$. Patch-level visual cues operate on a coarser granularity by dividing the image into fixed-size non-overlapping patches. Following the preprocessing schemes of mainstream VLMs, we adopt patches of size $h \times w$ pixels. Given an image with height $H$ and width $W$, we first ensure that both $H$ and $W$ are integer multiples of $h$ and $w$, enabling even partitioning into patches. For a pixel with absolute coordinates $(x, y)$, its corresponding patch coordinate $(r, c)$ is computed as:
\begin{equation}
r = \left\lfloor \frac{y}{h} \right\rfloor, \quad
c = \left\lfloor \frac{x}{w} \right\rfloor
\end{equation}
Thus, any pixel-level bounding box $[(x_1, y_1), (x_2, y_2)]$ can be converted to its patch-bbox representation by computing the top-left and bottom-right patch coordinates:
\begin{equation}
(r_1, c_1) = \left(\left\lfloor \frac{y_1}{h} \right\rfloor, \left\lfloor \frac{x_1}{w} \right\rfloor \right), \quad
(r_2, c_2) = \left(\left\lfloor \frac{y_2}{h} \right\rfloor, \left\lfloor \frac{x_2}{w} \right\rfloor \right)
\end{equation}
This two-dimensional patch coordinate $(r,c)$ serves as the patch ID for visual cue representation, which naturally aligns with VLM input tokenization and allows the model to attend to relevant image regions during reasoning.
In our experiments, the patch height and width ($h$ and $w$) are set to 28 to match the image loading format of Qwen-2.5-VL~\cite{bai2025qwen2}.

\subsection{Data Construction}
\label{Data Construction}

To fully exploit patch\mbox{-}bbox visual cues and enable the model to learn a robust interleaved visual--text reasoning paradigm, we develop a high\mbox{-}quality automated pipeline for constructing visual\mbox{-}cue--guided reasoning data, allowing large\mbox{-}scale generation of interleaved multimodal reasoning samples. As shown in Figure~\ref{fig:data pipeline}, the pipeline consists of the following stages:

\begin{figure}[t]
  \centering
  \includegraphics[width=0.85\textwidth]{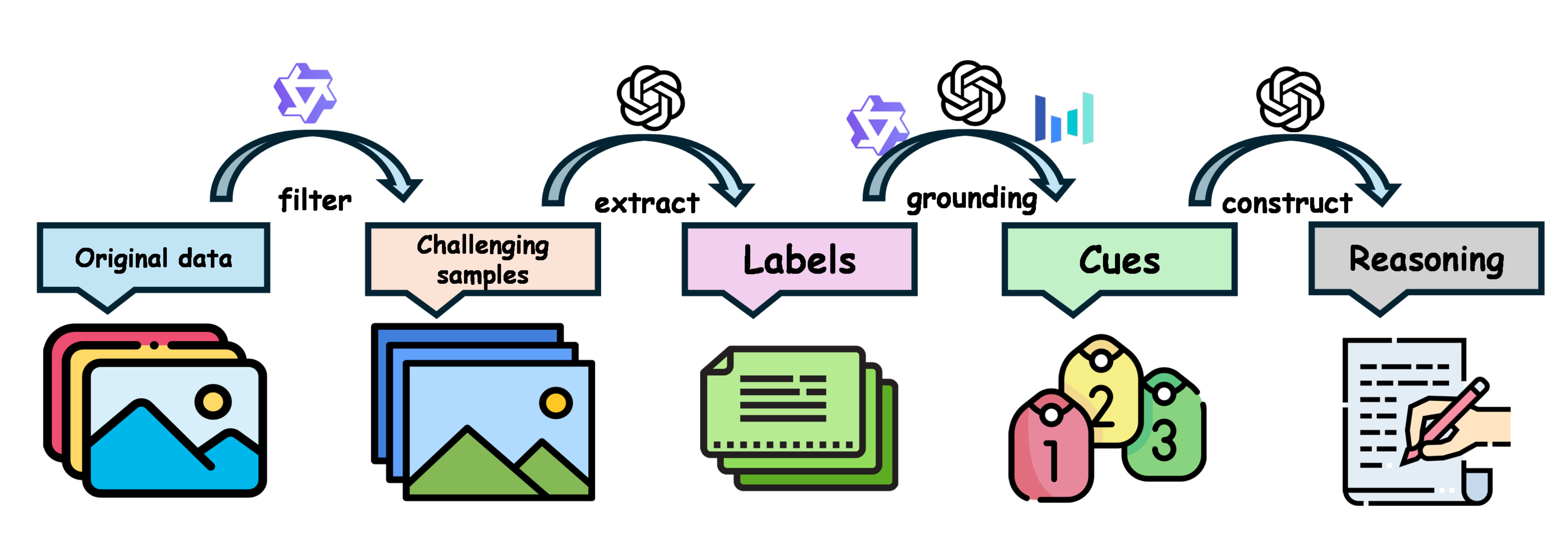}
  \caption{\textbf{Data Pipeline.} Starting from the collected original data, we filter to obtain challenging samples. Then extract and ground the key visual cues in the images, and finally construct new reasoning sequences based on these cues.} 
  \label{fig:data pipeline}
\end{figure}

\textbf{(1) Data Collection and Quality Filtering.}  
We first gather a variety of multimodal reasoning datasets, including CogCom \cite{qi2024cogcom}, DeepEyes \cite{zheng2025deepeyes}, Thyme \cite{zhang2025thyme}, and MINI-CoT \cite{chen2025mint}. To focus on challenging samples that can further improve reasoning capabilities, we filter the data using the base model Qwen2.5-VL-7B \cite{bai2025qwen2}, removing samples that the model can already answer correctly.

\textbf{(2) Visual Cue Extraction.}  
For the filtered samples, we use GPT-4o \cite{hurst2024gpt} to identify the critical visual regions needed to answer the questions, based on the image, question, and reference answers. The extracted regions are returned as structured cue labels.

\textbf{(3) Visual Cue Grounding.}  
To ensure precise localization, we retain model outputs as bbox coordinates and further validate them using three strong VLMs: GPT-4o \cite{hurst2024gpt}, Qwen2.5-VL-72B \cite{bai2025qwen2}, and Seed1.5-VL \cite{seed2025seed1_5vl}. We compute the IoU of the same cue labels across models, discarding samples where any pair falls below a threshold. Only samples with consistent and accurate localization across all three models are retained, and the bounding boxes are finally converted into patch-level representations.

\textbf{(4) Reasoning Construction.}  
Based on the original image question-answer pairs and the verified cue labels, GPT-4o \cite{hurst2024gpt} organizes the patch-level cues into complete reasoning sequences, which are then used for model training and optimization.

\begin{figure}[t]
  \centering
  \includegraphics[width=0.85\textwidth]{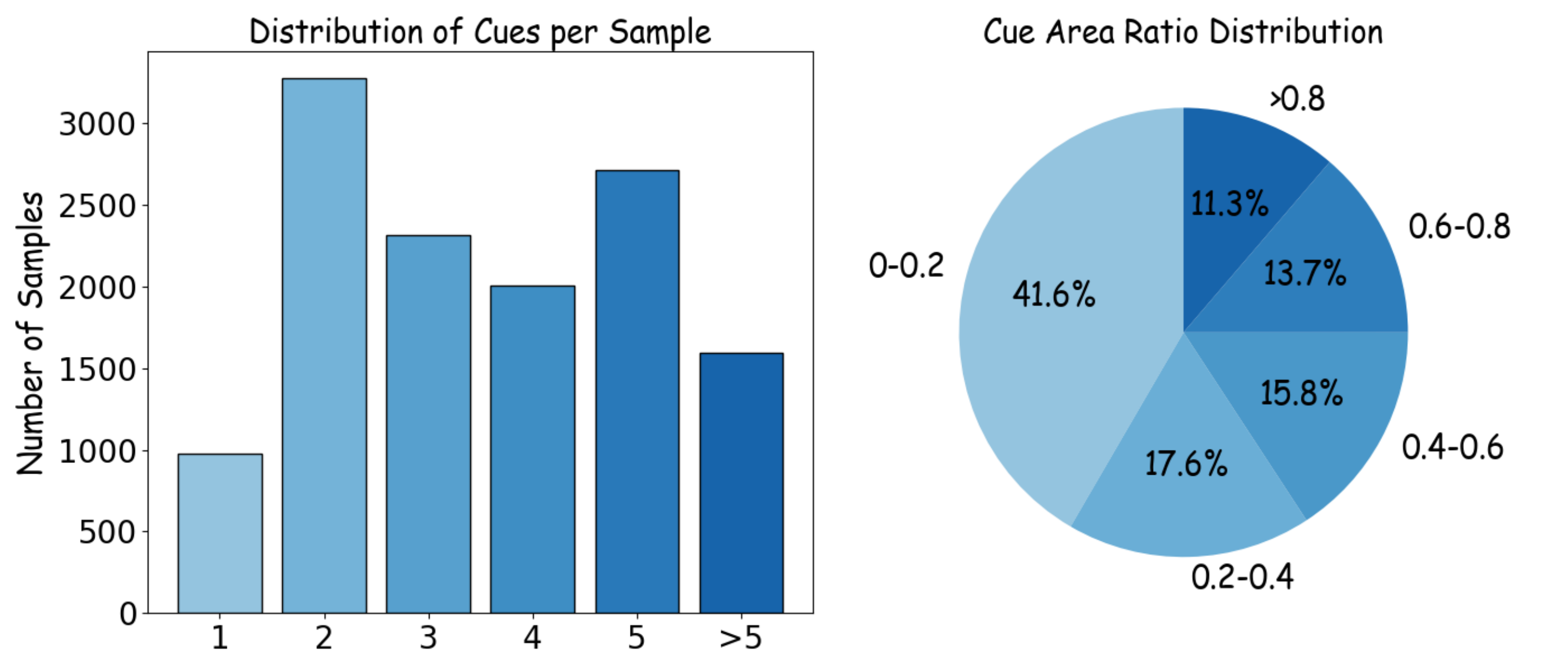}
  \caption{\textbf{Data Distribution.} In the left figure, we show the distribution of the number of cues per sample, where most cue data are concentrated between 2 and 5 cues; in the right figure, we show the distribution of the proportion of cue regions, with the majority of samples having cue regions occupying less than 40\% of the image.} 
  \label{fig:data_dura}
\end{figure}

Finally, in Figure~\ref{fig:data_dura}, we present the distribution of the cue data we constructed, including the distribution of the number of cues per sample and the distribution of the proportion of cue regions.

\subsection{Training Paradigm}
\label{sec: traning}
\paragraph{\textbf{Cold-start Initialization.}}
We employ the patch-bbox cue data to perform SFT as a cold-start initialization, ensuring that the model acquires the ability to generate reasoning sequences guided by patch-level visual cues. To further enhance the model’s generalization capability and enable it to handle both scenarios suitable for cue-based reasoning and those that are not, we incorporate a portion of general multimodal SFT training data \cite{mathew2021docvqa, lindstrom2022clevr, schwenk2022okvqa, kazemi2023geomverse} during this stage. In total, we select 12K patch-cue samples and 12K general QA samples for mixed SFT training, balancing cue-specific learning with broader multimodal reasoning ability.

\paragraph{\textbf{Reinforcement Learning.}}
To further enhance the model’s capability to autonomously generate visual cues, ensure their accuracy, and improve the alignment between the reasoning process and image content, we apply RL on the cold-started model using the GRPO algorithm \cite{shao2024deepseekmath}. To maximize the effectiveness of GRPO training, we first refine the training data by having the cold-started model perform multiple reasoning attempts on the candidate samples. Samples that the model consistently answers correctly or fails to answer are excluded, resulting in a curated set of 15K samples for GRPO training. GRPO then performs policy gradient optimization within each sample group, enabling the model to efficiently produce more diverse and richer reasoning sequences. The effectiveness of this approach largely depends on the design of the reward function, which in our framework consists of the following components:

$\bullet$ \textbf{Accuracy Reward:} The accuracy reward evaluates the model’s final output and is denoted as $R_{\text{acc}}$. It is computed by comparing the final answer extracted from the model’s reasoning process with the ground-truth answer. If the model’s final answer matches the ground-truth, $R_{\text{acc}}$ is set to 1; otherwise, it is set to 0.

$\bullet$ \textbf{Format Reward:} The model receives a reward of 1, denoted as $R_{\text{format}}$, if its output follows the required structured format, where the reasoning process, visual cues, and final answer are correctly enclosed within the \texttt{<think></think>}, \texttt{<cue></cue>}, and \texttt{<answer></answer>} tags, respectively.

$\bullet$ \textbf{Cue Reward:}
To evaluate the alignment between the model’s predicted visual cues and the GT cues, and to supervise the intermediate reasoning process, we design a patch-level $F_1$-based matching reward specifically tailored for the patch-form cues, denoted as $R_{\text{cue}}$. For each cue region, we construct the corresponding patch set:
\begin{equation}
\mathcal{S}(r_1,c_1,r_2,c_2) = \{(i,j) \mid r_1 \le i \le r_2,\, c_1 \le j \le c_2\},
\end{equation}
where $(r_1, c_1)$ and $(r_2, c_2)$ denote the top-left and bottom-right patch coordinates of a cue region. 
Given a predicted patch region $\mathcal{S}_p$ and a GT patch region $\mathcal{S}_g$, we define:
\begin{equation}
\text{TP} = |\mathcal{S}_p \cap \mathcal{S}_g|, \quad
\text{FP} = |\mathcal{S}_p \setminus \mathcal{S}_g|, \quad
\text{FN} = |\mathcal{S}_g \setminus \mathcal{S}_p|,
\end{equation}
With precision (Pre) and recall (Rec) computed as
\begin{equation}
Pre = \frac{\text{TP}}{\text{TP} + \text{FP}}, \quad
Rec = \frac{\text{TP}}{\text{TP} + \text{FN}},
\end{equation}
the patch-level $F_1$ score is then defined as
\begin{equation}
F_1 = \frac{2 \cdot Pre \cdot Rec}{Pre + Rec}.
\end{equation}
If the GT contains no cues and the model's reasoning output also contains no cues, $R_{\text{cue}}$ is set to 1. 
To ensure effective reasoning, if the number of predicted cues exceeds the number of GT cues, $R_{\text{cue}}$ is set to 0 to prevent the model from overproducing visual cues. 
When the number of predicted cues is less than or equal to the GT cues, we apply the Hungarian matching algorithm to find the optimal pairing between predicted and GT cues, ensuring a fair and structured evaluation of alignment.
 We construct the cost matrix:
\begin{equation}
C_{ij} = 1 - F_1(\mathcal{S}_p^i, \mathcal{S}_g^j).
\end{equation}
A matched pair $(i,j)$ is considered successful if
\begin{equation}
F_1(\mathcal{S}_p^i, \mathcal{S}_g^j) \ge \tau,
\end{equation}
where $\tau$ is a tunable hyperparameter controlling the minimum $F_1$ required for a successful match (default $\tau=0.5$).
Let $k$ denote the number of successful matches. The cue reward can then be defined as:
\begin{equation}
R_{\text{cue}} =
\frac{k}{|\mathcal{S}_g|}
\end{equation}
In summary, $R_{\text{cue}}$ can be uniformly expressed by the following formula:
\begin{equation}
R_{\text{cue}} =
\begin{cases}
1.0, & \text{if } |\mathcal{S}_p| = 0 \text{ and } |\mathcal{S}_g| = 0, \\
0, & \text{if } |\mathcal{S}_p| > |\mathcal{S}_g|, \\
\frac{k}{n_{\text{GT}}}, & \text{if } 0 < |\mathcal{S}_p| \le |\mathcal{S}_g|,
\end{cases}
\end{equation}
The final reward formulation is shown in Equation~\eqref{eq:final_reward}:

\begin{equation}
\label{eq:final_reward}
R = R_{\text{acc}} + R_{\text{format}} +R_{\text{cue}}
\end{equation}

\section{Experiment}

\subsection{Implementation Details}
\paragraph{\textbf{Test Benchmark.}} 
To thoroughly validate the effectiveness and generalization of PatchCue, evaluations were conducted on benchmarks covering diverse task dimensions. General question answering benchmarks include MMVet \cite{yu2023mm}, RealWorldQA \cite{xaiGrok1.5V}, MMStar \cite{chen2024we}, HallusionBench \cite{guan2024hallusionbench}, MMBench \cite{liu2024mmbench}, and MMVP \cite{tong2024eyes}. OCR-based document and chart understanding benchmarks include TextVQA \cite{singh2019towards}, AI2D \cite{kembhavi2016diagram}, OCRBench \cite{liu2024ocrbench}, and ChartQA \cite{masry2022chartqa}. Complex multimodal reasoning benchmarks include MMMU \cite{yue2024mmmu}, MathVista Mini \cite{lu2023mathvista}, and MathVision \cite{wang2024measuring}. Perception and counting benchmarks include BLINK \cite{fu2024blink} and CountBench \cite{paiss2023teaching}. High-resolution image perception benchmarks include HR-Bench4K \cite{wang2025divide}, HR-Bench8K \cite{wang2025divide}, and V* \cite{wu2024v}. This comprehensive setup ensures that PatchCue’s effectiveness is validated across general understanding, reasoning, perception, and high-resolution visual domains.

\begin{table*}[ht]
\caption{\textbf{Main results across multiple benchmarks.} We evaluate the performance of various VLMs trained with our PatchCue paradigm, demonstrating consistent improvements over baseline models across diverse datasets. The notation ``+PC" indicates models trained with our patch-bbox visual cue data.}
\label{tab:patchcue_benchmark}
\centering
\scriptsize
\setlength{\tabcolsep}{4pt}
\begin{tabular}{p{1.65cm}|l|cc|cc|cc}
\toprule
\textbf{Category} & \textbf{Benchmark} & 
\textbf{\parbox{1.3cm}{\centering Qwen2.5-VL-3B}} & \textbf{\parbox{0.8cm}{\centering +PC}} &
\textbf{\parbox{1.3cm}{\centering Qwen2.5-VL-7B}} & \textbf{\parbox{0.8cm}{\centering +PC}} &
\textbf{\parbox{1.0cm}{\centering MiMo-VL-7B}} & \textbf{\parbox{0.8cm}{\centering +PC}} \\
\midrule
\multirow{6}{*}{\parbox{2cm}{\raggedright General visual question answering}} 
& HallusionBench & 46.3 & 47.5 & 52.9 & 53.5 & 52.3 & 53.7 \\
& MMVet & 60.0 & 63.2 & 69.7 & 74.2 & 72.2 & 75.8 \\
& MMBench& 79.1 & 79.1 & 82.2 & 82.5 & 83.2 & 83.8 \\
& MMStar & 55.9 & 56.8 & 63.9 & 66.2 & 67.2 & 67.4 \\
& MMVP & 70.7 & 70.3 & 77.7 & 79.3 & 72.7 & 74.3 \\
& RealWorldQA & 67.5 & 68.0 & 68.5 & 69.3 &73.3 & 77.5 \\
\midrule
\multirow{4}{*}{\parbox{2cm}{\raggedright Document \& chart understanding}}
& TextVQA  & 79.3 & 84.8 & 84.9 & 87.4 & 81.2 & 84.3 \\
& AI2D  & 81.6 & 81.4 & 83.9 & 84.7 & 83.2 & 85.2 \\
& ChartQA  & 84.0 & 83.8 & 87.3 & 88.1 & 84.4 & 85.9 \\
& OCRBench & 79.7 & 81.5 & 88.8 & 91.1 & 82.9 & 84.8 \\
\midrule
\multirow{3}{*}{\parbox{2cm}{\raggedright Multimodal reasoning}}
& MathVision & 21.2 & 23.3 & 25.1 & 27.8 & 57.9 & 57.0 \\
& MathVista mini  & 62.3 & 63.3 & 68.2 & 69.6 & 81.8 & 80.8 \\
& MMMU  & 53.1 & 54.0 & 52.8 & 55.8 & 64.6 & 62.6 \\
\midrule
\multirow{2}{*}{\parbox{2cm}{\raggedright Perception / counting}}
& BLINK & 47.6 & 48.9 & 56.4 & 56.6 & 62.5 & 62.6 \\
& CountBench  & 77.8 & 78.8 & 89.3 & 89.9 & 87.0 & 89.3 \\
\midrule
\multirow{3}{*}{\parbox{2cm}{\raggedright High-res perception}}
& HR-Bench4K  & 66.3 & 65.0 &68.8 & 72.3 &75.2 & 75.4 \\
& HR-Bench8K  & 63.5 & 63.7 &65.3 & 69.6 &70.6 & 73.8 \\
& V*  & 75.4 & 73.8 &76.4 & 79.7 & 80.6 & 85.3 \\
\midrule
\rowcolor{gray!15}
\parbox[c][2em][c]{1.2cm}{Average} &
\parbox[c][2em][c]{1.0cm}{avg} &
\parbox[c][2em][c]{0.9cm}{\centering 65.0} &
\parbox[c][2em][c]{0.8cm}{\centering 66.1 \\ (+1.1)} &
\parbox[c][2em][c]{0.9cm}{\centering 70.1} &
\parbox[c][2em][c]{0.8cm}{\centering 72.1 \\ (+2.0)} &
\parbox[c][2em][c]{0.9cm}{\centering 73.9} &
\parbox[c][2em][c]{0.8cm}{\centering 75.4 \\ (+1.5)} \\
\bottomrule
\end{tabular}
\end{table*}

\paragraph{\textbf{Training and Inference Setups.} } 
All of our training tasks were implemented using the MS-Swift \cite{zhao2024swiftascalablelightweightinfrastructure} framework, and all evaluation tasks were conducted using the VLMEvalKit \cite{duan2024vlmevalkit} framework. The experiments were performed on 32 NVIDIA H20 GPUs, each with 96GB of memory.
 
\begin{table*}[t]
\caption{\textbf{Performance comparison across different forms of visual cues.} 
We compare the impact of different visual cue formats on model performance under the same data scale and reasoning paradigm, 
where ``Baseline” denotes the original results of Qwen2.5-VL 7B. The other columns show the results after SFT training using the cue data corresponding to each representation type.}
\label{tab:patch_formats}
\centering
\scriptsize
\setlength{\tabcolsep}{5.5pt}
\begin{tabular}{p{1.8cm}|l|ccccccc}
\toprule
\textbf{Category} & \textbf{Benchmark} &
\textbf{\parbox{1.0cm}{\centering Baseline}} & 
\textbf{\parbox{0.8cm}{\centering Pixel-Bbox}} &
\textbf{\parbox{0.8cm}{\centering Pixel-Point}} &
\textbf{\parbox{0.8cm}{\centering Patch-Bbox}} &
\textbf{\parbox{0.8cm}{\centering Patch-Point}} &
\textbf{\parbox{0.8cm}{\centering Labels}} \\
\midrule

\multirow{6}{*}{\parbox{2cm}{\raggedright General visual question answering}}
& HallusionBench  & 52.9 & 51.5 & 53.0 & 52.5 & 53.1 & 52.9 \\
& MMVet          & 69.7 & 66.0 & 65.3 & 70.4 & 65.3 & 63.3 \\
& MMBench         & 82.2 & 81.2 & 80.0 & 82.1 & 80.1 & 78.6 \\
& MMStar         & 63.9 & 64.9 & 64.8 & 65.6 & 64.7 & 64.6 \\
& MMVP          & 77.7 & 78.1 & 77.3 & 78.7 & 78.0 & 77.6 \\
& RealWorldQA     & 68.5 & 69.3 & 69.3 & 69.3 & 69.9 & 69.9 \\
\midrule

\multirow{4}{*}{\parbox{2cm}{\raggedright Document \& chart understanding}}
& TextVQA   & 84.9 & 85.1 & 85.0 & 86.8 & 85.0 & 84.4 \\
& AI2D      & 83.9 & 84.4 & 83.6 & 84.6 & 84.1 & 84.4 \\
& ChartQA   & 87.3 & 87.2 & 87.8 & 87.9 & 87.9 & 88.2 \\
& OCRBench  & 88.8 & 91.0 & 91.3 & 91.3 & 91.2 & 90.0 \\
\midrule

\multirow{3}{*}{\parbox{2cm}{\raggedright Multimodal reasoning}}
& MathVision      & 25.1 & 26.7 & 29.7 & 27.1 & 28.3 & 27.8 \\
& MathVista mini  & 68.2 & 69.6 & 68.7 & 70.1 & 68.1 & 69.0 \\
& MMMU           & 52.8 & 51.3 & 50.3 & 55.3 & 50.1 & 51.1 \\
\midrule

\multirow{2}{*}{\parbox{2cm}{\raggedright Perception / counting}}
& BLINK       & 56.4 & 55.7 & 55.4 & 57.2 & 55.1 & 56.8 \\
& CountBench  & 89.3 & 84.7 & 85.0 & 87.6 & 80.9 & 83.2 \\
\midrule

\multirow{3}{*}{\parbox{2cm}{\raggedright High-res perception}}
& HR-Bench4K  & 68.8 & 72.4 & 71.2 & 72.3 & 71.8 & 71.5 \\
& HR-Bench8K  & 65.3 & 68.5 & 69.7 & 69.9 & 69.0 & 69.1 \\
& V*          & 76.4 & 79.0 & 79.6 & 79.7 & 79.6 & 79.5 \\
\midrule

\rowcolor{gray!15}
Average & avg & 70.1 & 70.4 & 70.4 & 71.6 & 70.4 & 70.1 \\
\bottomrule
\end{tabular}
\end{table*}

\subsection{Main Results}
Our main experimental results are summarized in Table~\ref{tab:patchcue_benchmark}. To comprehensively evaluate the effectiveness of the PatchCue data and training methodology, and to assess its adaptability across different model sizes and architectures, we conducted experiments on three VLMs: Qwen2.5-VL-3B \cite{bai2025qwen2}, Qwen2.5-VL-7B \cite{bai2025qwen2}, and MiMo-VL-7B \cite{coreteam2025mimovltechnicalreport}. All models were trained using our patch-cue data through a two-stage process, consisting of SFT followed by RL. As shown in the table, all models consistently demonstrate performance gains across multiple benchmarks compared with their original versions. For instance, Qwen2.5-VL-7B achieves an improvement of 2.3 points on MMStar \cite{chen2024we}, confirming the effectiveness of our approach. The consistent improvements across different architectures and model scales further validate that the cue-based interleaved reasoning paradigm serves as a general framework, providing universal benefits to various VLMs. Meanwhile, the lightweight 3B model may exhibit relatively smaller performance gains due to its weaker CoT reasoning capability.

\begin{table}[ht]
\centering
\caption{\textbf{Performance comparison under different training data setups.} We evaluate the impact of training data composition on model performance. ``Baseline” denotes the original Qwen2.5-VL-7B results. Other rows indicate training on different combinations of general (Gen) and patch-bbox cue (Cue) data, with ratios specified as Gen:Cue.}
\label{tab:dataset}
\setlength{\tabcolsep}{3pt} 
\renewcommand{\arraystretch}{1.25}

\begin{tabular}{
    l|cccc
}
\toprule
{\small \textbf{Training Setup}} & \textbf{AI2D\cite{kembhavi2016diagram}} & \textbf{ChartQA\cite{masry2022chartqa}} & \textbf{MMStar\cite{chen2024we}} & \textbf{MMVP\cite{tong2024eyes}} \\
\midrule
Baseline &83.9  &87.3  &63.9  &77.7 \\
Gen Only (1:0) &83.5  &87.5  &65.3  &76.7\\
Hybrid (2:1) &84.6  &87.5  &64.6 &78.0 \\
Hybrid (1:1) &84.6  &87.9  &65.6 &78.7\\
Hybrid (1:2) &85.1  &87.4  &64.9 &71.0  \\
Cue Only (0:1)  &80.8  &86.6  &60.3 &68.0 \\
\bottomrule
\end{tabular}
\end{table}

\subsection{Analysis}
\paragraph{\textbf{Impact of different cue formats on performance.}}
We systematically investigate the impact of different visual cue representations on model performance during training. Specifically, we adopt Qwen2.5-VL-7B \cite{bai2025qwen2} as the base model and transform all \textbf{patch-bbox} cues used in the SFT stage into several alternative formats: (1) pixel-level bounding boxes represented by relative pixel coordinates (\textbf{pixel-bbox}), (2) single-pixel location cues (\textbf{pixel-point}), (3) patch-level cues represented by the coordinates of the central patch region (\textbf{patch-point}), and (4) a text-only variant that removes visual cues while retaining textual labels (\textbf{labels}). Throughout this process, the overall dataset size and content remain unchanged, with only the cue representation format being modified, ensuring a fair comparison. We then retrain the model with each cue type and evaluate its performance across multiple benchmarks to assess the influence of cue design. As shown in Table~\ref{tab:patch_formats}, under identical data scales and training paradigms, the models exhibit varying performance depending on the cue format. Notably, the patch-bbox representation consistently achieves the largest overall improvement, highlighting its effectiveness and generalizability as a superior visual cue for multimodal reasoning tasks.

\paragraph{\textbf{Ablation study on data composition.}}
During the SFT training stage, we incorporate a portion of non-cue general data along with our patch-bbox cue data for mixed training. To evaluate the contribution of cue data, we conduct a data ratio ablation study, where the total amount of SFT training data is kept constant while varying the proportion of visual cue data and general non-cue data. We train Qwen2.5-VL-7B \cite{bai2025qwen2} under different ratio settings and compare the resulting model performances. As shown in Table~\ref{tab:dataset}, models trained solely on non-cue data achieve only marginal improvements, indicating that cue-based data effectively enhances the model’s perceptual reasoning capabilities. However, using only cue data leads to performance drops on certain benchmarks, which we attribute to the reduced output diversity and instruction-following ability caused by SFT training exclusively on cue data. This suggests that while cue data is crucial for improving visual reasoning, an appropriate balance with general data is necessary to maintain overall model robustness.

\begin{table}[ht]
\centering
\caption{\textbf{Ablation of cue reward.} We compare the performance differences between RL training with and without the use of $R_{\text{cue}}$. ``Baseline” denotes the original Qwen2.5-VL-7B results, ``+SFT” represents the results after SFT training, ``+RL (w/o $R_{\text{cue}}$)” indicates RL training based on the SFT model without applying $R_{\text{cue}}$, and ``+RL” denotes RL training with the incorporation of $R_{\text{cue}}$.}
\label{tab:reward}
\setlength{\tabcolsep}{2pt} 
\renewcommand{\arraystretch}{1.25}

\begin{tabular}{l|cccc}
\toprule
{\footnotesize \textbf{Model}} & {\footnotesize \textbf{AI2D\cite{kembhavi2016diagram}}} & {\footnotesize \textbf{ChartQA\cite{masry2022chartqa}}} & {\footnotesize \textbf{MMStar\cite{chen2024we}}} & {\footnotesize \textbf{MMVP\cite{tong2024eyes}}}
 \\
\midrule
Baseline & 83.9 & 87.3 & 65.3 & 77.7 \\
+SFT & 84.6 & 87.9 & 65.6 & 78.7 \\
+RL(w/o $R_{\text{cue}}$) &83.9  &87.5  &65.7  &78.7  \\
+RL & 84.7 & 88.1 & 66.2 & 79.3 \\
\bottomrule
\end{tabular}
\end{table}

\paragraph{\textbf{Ablation of cue reward.}}
During the GRPO training stage, we introduce a novel process-level reward function specifically designed for patch-bbox cues. Table~\ref{tab:reward} presents a comparison between models trained with and without this cue-specific reward. The results indicate that incorporating the cue reward not only yields more substantial performance gains but also leads to more stable and consistent training dynamics. These findings highlight the effectiveness of the proposed reward function in guiding the model to better leverage visual cues, ultimately enhancing reasoning quality and overall performance.

\begin{table}[h]
\centering
\caption{Performance comparison of different methods on multimodal benchmarks. We sample $\sim$12K instances from VisualCoT, CogCom, and MINI-CoT, and fine-tune the same backbone (Qwen2.5-VL-7B) with an identical SFT protocol for a fair comparison across methods with different original settings.}
\label{tab:benchmark_comparison}
\begin{tabular}{lcccc}
\toprule
\textbf{Method} & \textbf{AI2D\cite{kembhavi2016diagram}} & \textbf{ChartQA\cite{masry2022chartqa}} & \textbf{MMStar\cite{chen2024we}} & \textbf{MMVP\cite{tong2024eyes}} \\
\midrule
VisualCoT \cite{shao2024visual}  &  83.9  &  85.0  &  63.8  &  75.2  \\
CogCom \cite{qi2024cogcom}    &  84.6  &  85.6  &  64.0  &  76.0  \\
MINI-CoT \cite{chen2025mint}   &  84.1  &  87.9  &  64.8  &  77.7  \\
PatchCue   &  \textbf{84.7}  &  \textbf{88.1}  &  \textbf{66.2}  &  \textbf{79.3}  \\
\bottomrule
\end{tabular}
\end{table}
\paragraph{\textbf{Comparison with other methods.}} To more clearly demonstrate the effectiveness of PatchCue, we compare it with several other methods for incorporating visual cues, as shown in Table 1. Since the experimental details of different methods vary, for example, CogCom~\cite{qi2024cogcom} uses external visual tool calls and MINI-CoT~\cite{chen2025mint} modifies the model backbone by adding extra encoding layers, we conduct a more direct comparison by training all methods with the same backbone, Qwen2.5-VL-7B~\cite{bai2025qwen2}. For each method, we randomly sample about 12K training instances from their cue data for SFT training. The results in the table show that, under the same experimental settings and data size, PatchCue provides the most significant performance improvement.

\begin{figure}[ht]
  \centering
  \includegraphics[width=1.\textwidth]{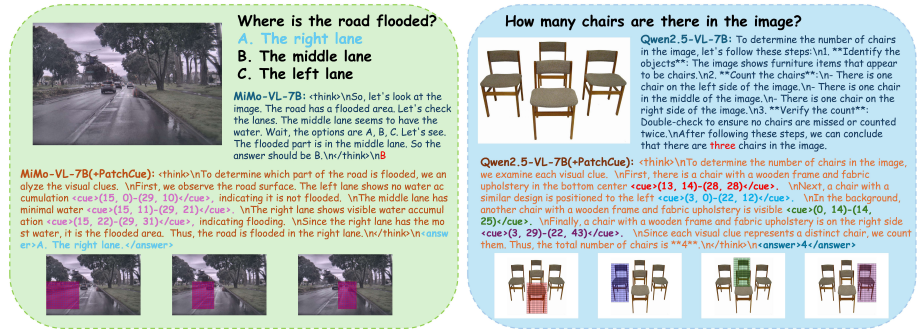}
  \caption{\textbf{Case Study.} We compare the model’s outputs before and after PatchCue training. After training, the model can generate visual cues during reasoning, improving both its perception and the interpretability of its reasoning process.} 
  \label{fig:vis_case}
\end{figure}

\paragraph{\textbf{Case Study.}}
We illustrate the reasoning outputs of MiMo-VL-7B \cite{coreteam2025mimovltechnicalreport} and Qwen2.5-VL-7B \cite{bai2025qwen2}, highlighting the differences before and after training with PatchCue in Figure~\ref{fig:vis_case}. After training with patch-bbox cue data, the models gain the ability to explicitly generate visual cues throughout the reasoning process. This improvement not only strengthens their multimodal understanding and reasoning performance but also enhances the transparency and interpretability of their reasoning chains, making it easier to verify how visual information contributes to their conclusions.

\subsection{Discussion}
Based on our proposed framework and experimental findings, several additional insights and methodological discussions can be derived.

\begin{center}
\begin{tcolorbox}[
    colback=blue!5!white,     
    colframe=blue!40!black,   
    boxrule=0.5mm,
    width=0.8\textwidth,
    arc=3mm,
    auto outer arc,
    breakable,
    top = 1pt,
    bottom = 2pt
]
\textbf{Findings 1:} Simulating human visual perception requires more sophisticated reasoning paradigms.
\end{tcolorbox}
\end{center}
Our experiments indicate that cue-output reasoning can help models address specific perceptual challenges, but relying solely on cue-based training data may degrade performance. In reality, humans do not rely on a single reasoning paradigm during perception; depending on the context, they flexibly combine visual cues, background knowledge, and experiential reasoning to interpret complex information. Similarly, in complex and diverse application scenarios, VLMs need the ability to adaptively switch or integrate multiple reasoning strategies to effectively handle more challenging tasks.

\begin{center}
\begin{tcolorbox}[
    colback=blue!5!white,     
    colframe=blue!40!black,   
    boxrule=0.5mm,
    width=0.8\textwidth,
    arc=3mm,
    auto outer arc,
    breakable,
    top = 1pt,
    bottom = 2pt
]
\textbf{Findings 2:} Models require more flexible forms of visual cues to achieve true “think with images” capabilities.
\end{tcolorbox}
\end{center}
We explored a new form of visual cue representation and verified its effectiveness. However, under our patch-bbox visual cue framework, some base models show suboptimal performance on certain tasks (Table~\ref{tab:patchcue_benchmark}). In comparison, point-based cues at either the pixel or patch level achieve better results on benchmarks designed for mathematical and geometric reasoning, such as MathVision \cite{wang2024measuring} (Table~\ref{tab:patch_formats}). This indicates that human visual cue perception relies on more complex and diverse cues, and more general and flexible visual cues may be needed to fully realize “think with images” capabilities.

\begin{center}
\begin{tcolorbox}[
    colback=blue!5!white,     
    colframe=blue!40!black,   
    boxrule=0.5mm,
    width=0.8\textwidth,
    arc=3mm,
    auto outer arc,
    breakable
]
\textbf{Findings 3:} Process rewards can play an effective role in perceptual reasoning.
\end{tcolorbox}
\end{center}
In Table~\ref{tab:reward}, we present a comparison of GRPO training results with and without the incorporation of process-level visual cue rewards. The results demonstrate that introducing such rewards, particularly under task-specific settings, can effectively enhance the model’s performance by guiding the intermediate perceptual reasoning process and stabilizing training outcomes.

\section{Conclusion}
We propose PatchCue, a patch-bbox-based visual cue paradigm that divides images into patches and encodes cues at the patch level, aligning with human perceptual habits and the patch-tokenized structure of modern VLMs. Using a two-stage training strategy combining SFT and process-supervised RL, PatchCue enables models to generate and leverage visual cues more effectively during interleaved visual-text reasoning. Experiments across multiple VLMs and benchmarks show that patch-bbox cues consistently improve performance, indicating that well-designed visual cue representations can enhance multimodal reasoning and guide future cognitively aligned VLM research.


%
%
\clearpage
\bibliographystyle{splncs04}
\bibliography{main}

\clearpage

\section{Supplementary}
\subsection{Theoretical Details of GRPO}
The complete policy optimization objective for GRPO training is as follows:
\begin{equation}
  \resizebox{\linewidth}{!}{$%
    \begin{aligned}
      \mathcal{J}_{\text{GRPO}}(\theta)
      & =\mathbb{E}\!\left[q \sim P(Q),\{o_i\}_{i=1}^{G} \sim \pi_{\theta_{\text{old}}}(O \mid q)\right] \\
      &\quad \frac{1}{G}\sum_{i=1}^{G}\frac{1}{|o_i|}\sum_{t=1}^{|o_i|}\!\Bigl\{
          \min\!\Bigl[
            \frac{\pi_{\theta}(o_{i,t}\!\mid\! q, o_{i,<t})}{\pi_{\theta_{\text{old}}}(o_{i,t}\!\mid\! q, o_{i,<t})}
            A_{i,t}, \\
      & 
            \operatorname{clip}\!\Bigl(
              \frac{\pi_{\theta}(o_{i,t}\!\mid\! q, o_{i,<t})}{\pi_{\theta_{\text{old}}}(o_{i,t}\!\mid\! q, o_{i,<t})},
              1-\varepsilon, 1+\varepsilon
            \Bigr) A_{i,t}
          \Bigr] -\beta\,\mathbb{D}_{\mathrm{KL}}\!\bigl[\pi_{\theta}\,\|\,\pi_{\mathrm{ref}}\bigr]\Bigr\},
    \end{aligned}$}%
\end{equation}
where for each input question \(q\) sampled from the distribution \(P(Q)\), the rollout module generates a group of trajectories \(\{o_i\}_{i=1}^{G}\) from the old policy \(\pi_{\theta_{\text{old}}}\) through interaction with the external environment. The term \(A_{i,t}\) represents the advantage at step \(t\) of trajectory \(i\), computed based on the relative rewards of outputs within the group. The reward function consists of the three components introduced in the main text: Accuracy Reward, Format Reward, and Cue Reward.

\subsection{Training Setting Details}
\label{sec:Training Setting Details.}

During the cold-start phase, we perform full-model fine-tuning, with the relevant hyperparameters listed in the Table~\ref{tab:train} below. All other parameters not listed are kept consistent with Swift’s official documentation.
\begin{table}[ht]
\centering
\caption{Hyperparameters for SFT Training.}
\setlength{\tabcolsep}{12pt} 
\renewcommand{\arraystretch}{1.2} 
\label{tab:train}
\begin{tabular}{c|c}
\toprule
Hyperparameter & Settings \\ \midrule
DeepSpeed Stage & 2 \\
Warmup Ratio & 0.05 \\
Trainable Module & LLM \\
Epoch & 1 \\
LR Schedule & cosine \\
Learning Rate & 1e-5 \\
Max Pixels & 1003520 \\
Torch Dtype & bfloat16 \\
Batch Size & 128 \\ \bottomrule
\end{tabular}
\end{table}

During the RL phase, we train the model using the GRPO algorithm, with the corresponding hyperparameters listed in Table~\ref{tab:grpo_train}. All other parameters not listed are kept consistent with Swift’s official documentation.

\begin{table}[ht]
\centering
\caption{Hyperparameters for GRPO Training}
\label{tab:grpo_train}
\setlength{\tabcolsep}{12pt}
\renewcommand{\arraystretch}{1.2}
\begin{tabular}{@{}c|c@{}}
\toprule
Hyperparameter & Settings \\ \midrule
Beta & 0.001 \\
Torch Dtype & bfloat16 \\
Learning Rate & 1e-6 \\
Warmup Ratio & 0.05 \\
Num Generations & 8\\
Epoch & 3 \\
DeepSpeed Stage  & 3  \\
Temperature & 1.0 \\
Top-p & 1.0 \\
Top-k & 80 \\
Repetition Penalty & 1.1 \\
Epsilon & 0.1 \\
Batch Size & 128 \\
Max Completion Length & 2048 \\
\bottomrule
\end{tabular}
\end{table}

\subsection{More Cases}

\begin{figure*}[t]
  \centering
  \includegraphics[width=\textwidth]{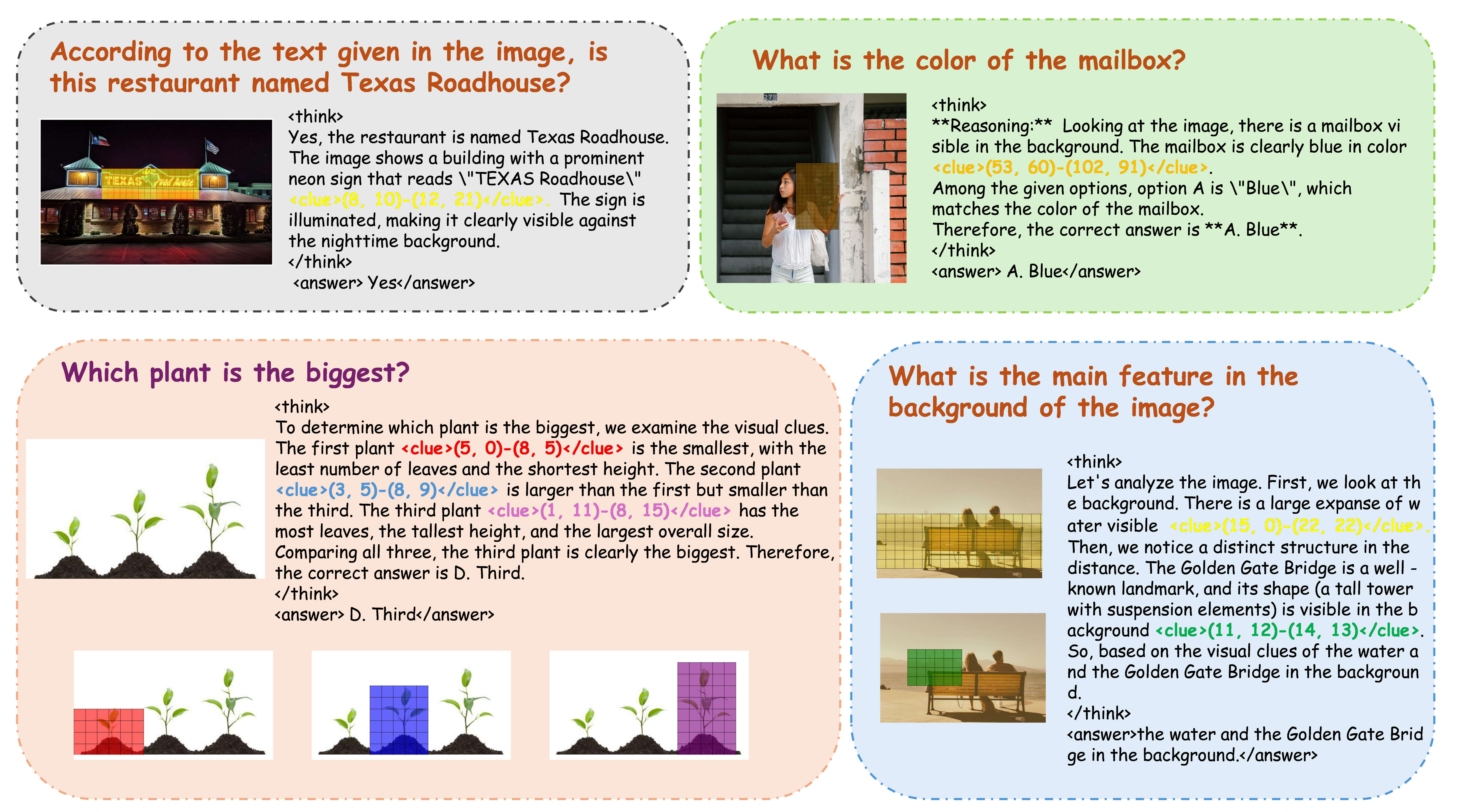}
  \caption{Successful examples where visual cues effectively guide the model to conduct accurate and interpretable reasoning.}
  \label{fig:case1}
\end{figure*}

We present additional qualitative examples in Figure~\ref{fig:case1}, which demonstrate how visual cues effectively guide the model toward task-relevant regions and facilitate more accurate and interpretable reasoning. These results further verify that incorporating cue information strengthens the model's ability to ground its reasoning in the visual content.

\begin{figure*}[t]
  \centering
  \includegraphics[width=\textwidth]{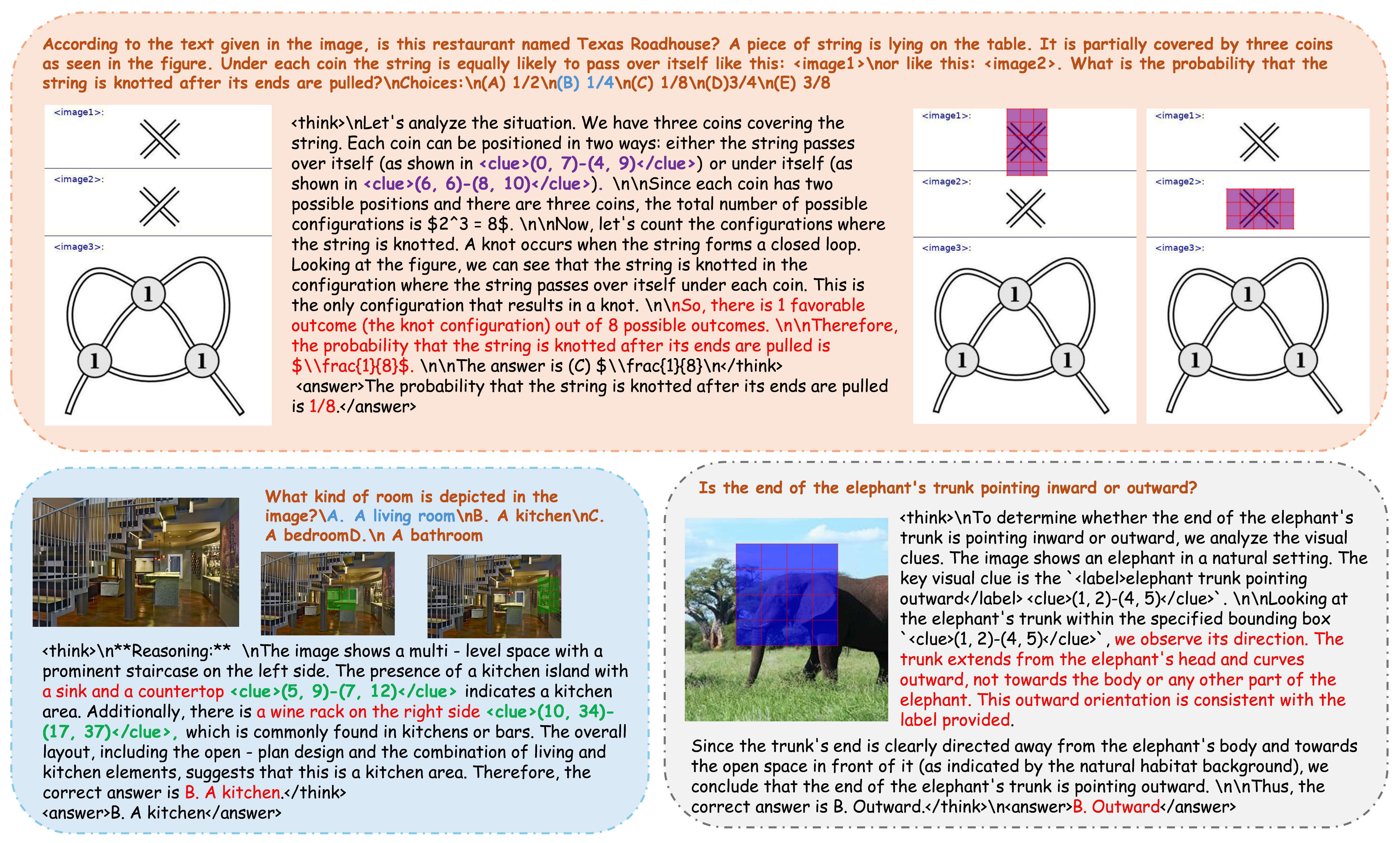}
  \caption{Common failure cases of cue-based reasoning, including flawed reasoning despite reasonable cues, incorrect cue localization, and localization deviations that mislead the final prediction.}
  \label{fig:bad_case}
\end{figure*}

In Figure~\ref{fig:bad_case}, we analyze several representative failure modes of cue-based reasoning:
\begin{itemize}
    \item The model identifies reasonably accurate cues or the cues themselves are not essential, yet the subsequent logical reasoning is incorrect.
    \item The model fails to locate the correct cues, resulting in its reasoning process being misdirected by inaccurate visual information.
    \item The model identifies the general cue region but with noticeable localization errors, ultimately leading to incorrect final predictions.
\end{itemize}

\subsection{Prompt Template}
We provide the reference prompts used in constructing the visual cue data, as illustrated in Figures~\ref{fig:prompt_cues_text}, \ref{fig:prompt_cues_bbox}, and \ref{fig:prompt_cot_with_cues_part2}.

\clearpage
\clearpage
\begin{figure*}[ht]
\centering
\begin{tcolorbox}[breakable, colback=gray!5!white, colframe=gray!75!black,
title=Cue Extraction Prompt, boxrule=0.5mm, width=\textwidth, arc=3mm, auto outer arc]
You will receive an image, a question related to the image, and the final answer to that question. Your task: \\[2mm]

\begin{itemize}
  \item Extract \textbf{the key visual cues} from the image that are directly related to answering the question.
  \item For each cue, provide a \textbf{textual description wrapped in \texttt{<label></label>}}, without any bounding box information.
  \item Output \textbf{only the most critical cues}, \textbf{one per line}, and \textbf{at most 5 cues}. \\
\end{itemize} 

\textbf{Example 1:} \\ 
Question: The members of the local garden club tallied the number of plants in each person's garden. How many gardens have at least 48 plants but fewer than 80 plants? \\ 
Final answer: 6 \\ 
Extracted cues: \\
\texttt{<label>row with stem = 4</label>}   \quad \texttt{<label>row with stem = 8</label>} \\[1mm]

\textbf{Example 2:} \\ 
Question: Who invented the heater in this image? \\ 
Final answer: franz san galli \\ 
Extracted cues: \texttt{<label>heater</label>} \\[1mm]

\textbf{Example 3:} \\ 
Question: How many flowers are there?\\
Choices: (A) 87, (B) 94, (C) 79 \\ 
Final answer: 87 \\ 
Extracted cues: \\
\texttt{<label>10 flowers in the first row</label>} \\
\texttt{<label>9 flowers in the first column</label>} \\
\texttt{<label>three empty spots in the last row</label>} \\[1mm]

\textbf{Now process the following:} \\ 
Question: \{question\} \\ 
Final answer: \{answer\} \\[1mm]

\textbf{Your extracted visual cues (text only, wrapped in \texttt{<label>}):}
\end{tcolorbox}
\caption{Cue Extraction Prompt}
\label{fig:prompt_cues_text}
\end{figure*}

\begin{figure*}[ht]
\centering
\begin{tcolorbox}[breakable, colback=gray!5!white, colframe=gray!75!black,
title=Cue Grounding Prompt, boxrule=0.5mm, width=\textwidth, arc=3mm, auto outer arc]
You will receive an image and a list of textual visual cues extracted from the image. Your task: \\[2mm]

For each cue, provide the location in the image as a \textbf{normalized bounding box}, in the format: 
\texttt{<bbox>[x\_min, y\_min, x\_max, y\_max]</bbox>}. 
Output \textbf{one line per cue}, preserving the same order as the input cues. \\

\textbf{Example:} \\ 
Cues: \\
\texttt{<label>heater</label>} \quad \texttt{<label>light switch</label>} \\ 
Extracted bounding boxes: \\
\texttt{<bbox>[0.235, 0.345, 0.521, 0.876]</bbox>} \quad \texttt{<bbox>[0.120, 0.400, 0.300, 0.520]</bbox>} \\[1mm]

\textbf{Now process the following:} \\ 
Cues: {cues} \\[1mm]

\textbf{Your extracted bounding boxes (wrapped in \texttt{<bbox>}):}
\end{tcolorbox}
\caption{Cue Grounding Prompt}
\label{fig:prompt_cues_bbox}
\end{figure*}

\begin{figure*}[ht]
\centering
\begin{tcolorbox}[breakable, colback=gray!5!white, colframe=gray!75!black,
title=Reasoning Construction Prompt, boxrule=0.5mm,
width=\textwidth, arc=3mm, auto outer arc]

You will be given an image, a question related to the image, the correct answer, and one or more \textbf{key visual cues}. Your task is:

\begin{itemize}
  \item Generate a \textbf{complete and logical reasoning process} based on the given visual cues, leading to the correct answer.
  \item Each visual cue contains two components: a \texttt{<label>} describing the visual content, and a \texttt{<bbox>} specifying the corresponding bounding box.
  \item Every time you refer to a visual cue in your reasoning, you \textbf{must also include its corresponding \texttt{<bbox>}} using the \texttt{<bbox></bbox>} tag.
\end{itemize}

\textbf{Requirements:}  

The reasoning must be \textbf{clear, structured, and logically connected}.  
When using visual evidence in reasoning, always quote the exact bbox along with it.

---

\textbf{Example 1:} \\ 
\textbf{Question:} \\ 
\texttt{<image>} \\ 
How many flowers are there? \\
Choices: (A) 87, (B) 94, (C) 79 \\[1mm]

\textbf{Answer:} \\ 
(A) 87 \\

\end{tcolorbox}
\label{fig:prompt_cot_with_cues_part1}
\end{figure*}

\begin{figure*}[ht]
\centering
\begin{tcolorbox}[breakable, colback=gray!5!white, colframe=gray!75!black,
title=, boxrule=0.5mm,
width=\textwidth, arc=3mm, auto outer arc]

\textbf{Example 1 (continued) - Visual Cues + Reasoning:} \\ 
\textbf{Visual Cues:} \\ 
\texttt{<label>10 flowers in the first row</label>} \quad \texttt{<bbox>[0.0, 0.0, 0.909, 0.120]</bbox>} \\
\texttt{<label>9 flowers in the first column</label>} \quad \texttt{<bbox>[0.0, 0.0, 0.120, 0.916]</bbox>} \\
\texttt{<label>three empty spots in the last row</label>} \quad \texttt{<bbox>[0.621, 0.832, 0.910, 0.916]</bbox>} \\[1mm]

\textbf{Reasoning:} \\ 
Looking at the image, we can see that the flowers are arranged in a regular rectangular pattern. The first row has over ten flowers \texttt{<bbox>[0.0, 0.0, 0.909, 0.120]</bbox>}, meaning each row has over ten flowers. The first column has nine flowers \texttt{<bbox>[0.0, 0.0, 0.120, 0.916]</bbox>}, meaning each column has nine flowers. Furthermore, three positions in the last row are empty \texttt{<bbox>[0.621, 0.832, 0.910, 0.916]</bbox>}, so the final number of flowers is 9 × 10 - 3 = 87. The final answer is (A) 87.

---

\textbf{Example 2:} \\ 
\textbf{Question:} \\ 
 \texttt{<image>} \\ 
Who was the most valuable Costa Rican soccer player as of June 2018? (Answer with a single word.) \\[1mm]

\textbf{Answer:} \\ 
Keylor Navas \\[1mm]

\textbf{Visual Cues:} \\ 
\texttt{<label>the longest bar</label>} \quad \texttt{<bbox>[0.194, 0.062, 0.851, 0.141]</bbox>} \\
\texttt{<label>Keylor Navas</label>} \quad \texttt{<bbox>[0.107, 0.062, 0.194, 0.141]</bbox>} \\[1mm]

\textbf{Reasoning:} \\ 
The image is a bar chart showing the market value of Costa Rican soccer players.  
The longest bar \texttt{<bbox>[0.194, 0.062, 0.851, 0.141]</bbox>} represents the highest market value.  
To its left, the associated label is "Keylor Navas" \texttt{<bbox>[0.107, 0.062, 0.194, 0.141]</bbox>}.  
Therefore, the most valuable player is \textbf{Keylor Navas}.

---

\textbf{Now process the following:} \\ 
Question: {question} \\ 
Answer: {answer} \\ 
Visual Cues: {clues} \\[1mm]

\textbf{Your response:}

\end{tcolorbox}
\caption{Reasoning Construction Prompt.}
\label{fig:prompt_cot_with_cues_part2}
\end{figure*}

\end{document}